\documentclass[preprint,12pt]{elsarticle}
\biboptions{sort&compress}

\usepackage{amssymb}
\usepackage{amsmath}
\usepackage{amsfonts}
\usepackage{IEEEtrantools}
\usepackage{pdflscape}
 \usepackage{geometry}
\usepackage{algorithmic}
\usepackage{array}
\usepackage[caption=false,font=normalsize,labelfont=sf,textfont=sf]{subfig}
\usepackage{textcomp}
\usepackage{stfloats}
\usepackage{url}
\usepackage{verbatim}
\usepackage{algorithm}
\usepackage{multirow}
\usepackage{booktabs}  
\usepackage{subfig}
\usepackage{graphicx}  
\usepackage{booktabs} 
\usepackage{xfrac}
\usepackage{tabularx}
\usepackage{threeparttable}
\usepackage{adjustbox}
\usepackage{makecell}
\usepackage{xcolor}
\usepackage{epstopdf}
\usepackage{etoolbox}

\makeatletter
\def\ps@pprintTitle{%
  \let\@oddhead\@empty
  \let\@evenhead\@empty
  \def\@oddfoot{\hfil\thepage\hfil}
  \let\@evenfoot\@oddfoot
}
\makeatother

\begin{document}

\begin{frontmatter}



\title{Introducing Fractional Classification Loss for Robust Learning with Noisy Labels
}


\author[inst1]{Mert Can Kurucu\corref{cor1}}
\ead{kurucum@itu.edu.tr}

\affiliation[inst1]{organization={Faculty of Electrical and Electronics Engineering},
            addressline={Istanbul Technical University}, 
            city={Istanbul},
            postcode={34469}, 
            country={Türkiye}}

 \affiliation[inst2]{organization={AI and Intelligent Systems Laboratory},
 addressline={Istanbul Technical University}, 
 city={Istanbul},
  postcode={34469}, 
  country={Türkiye}}
\author[inst2]{Tufan Kumbasar}
\ead{kumbasart@itu.edu.tr}
\author[inst1]{Ibrahim Eksin}
\ead{eksin@itu.edu.tr}
\author[inst1]{M\"ujde G\"uzelkaya}
\ead{guzelkaya@itu.edu.tr}

\cortext[cor1]{Corresponding author}

\begin{abstract}
Robust loss functions are crucial for training deep neural networks in the presence of label noise, yet existing approaches require extensive, dataset-specific hyperparameter tuning. In this work, we introduce Fractional Classification Loss (FCL), an adaptive robust loss that automatically calibrates its robustness to label noise during training. Built within the active-passive loss framework, FCL employs the fractional derivative of the Cross-Entropy (CE) loss as its active component and the Mean Absolute Error (MAE) as its passive loss component. With this formulation, we demonstrate that the fractional derivative order \(\mu\) spans a family of loss functions that interpolate between MAE-like robustness and CE-like fast convergence. Furthermore, we integrate \(\mu\) into the gradient-based optimization as a learnable parameter and automatically adjust it to optimize the trade-off between robustness and convergence speed. We reveal that FCL's unique property establishes a critical trade-off that enables the stable learning of $\mu$: lower log penalties on difficult or mislabeled examples improve robustness but impose higher penalties on easy or clean data, reducing model confidence in them. Consequently, FCL can dynamically reshape its loss landscape to achieve effective classification performance under label noise. Extensive experiments on benchmark datasets show that FCL achieves state-of-the-art results without the need for manual hyperparameter tuning.
\end{abstract}



\begin{keyword}
deep learning \sep adaptive loss \sep robust loss function \sep fractional calculus \sep label noise \sep multiclass classification 
\end{keyword}

\end{frontmatter}


\section{Introduction}
Training Deep Neural Networks (DNNs) in the presence of noisy labels remains a significant and longstanding challenge, with considerable practical implications \cite{natarajan2013learning, liu2023noise}. Label noise severely impacts model accuracy, emphasizing the importance of designing robust loss functions to ensure reliable performance. Recently, fractional calculus, a mathematical approach that extends differentiation and integration to non-integer orders, has demonstrated significant potential to boost the robustness and performance of DNN architectures \cite{wei2020generalization}, activation functions \cite{zamora2022fractional,kumar2024enhancing}, backpropagation algorithms \cite{bao2018fractional}, and loss functions \cite{kurucu2025fractional}. In particular, \cite{kurucu2025fractional} proposes an adaptive robust regression loss function that leverages fractional derivatives to dynamically reshape the loss landscape to balance the robustness and convergence speed. The fractional order parameter $\mu$ is learned adaptively, enabling the loss to respond to varying noise in the dataset effectively.

\subsection{Motivation and Related Work}
The current paradigm of machine learning is driven by large models trained on vast datasets. However, this reliance on large-scale data acquisition often prioritizes quantity over quality, leading to datasets that contain significant noise, which can degrade model performance and generalization \cite{tang2024advancing,liu2024towards}. In supervised classification tasks, data labeling, whether performed automatically, by humans, or through web scraping, produces large datasets that frequently contain mislabeling errors. These errors can result from human mistakes, limited expertise, or the error-prone nature of automated methods \cite{patrini2017making,an2025relative}. Such mislabels introduce various forms of label noise that negatively affect model training \cite{zhou2023asymmetric,li2025cltr}. 

Several strategies have been proposed to tackle learning with label noise, including label correction, sample selection, refined training methodologies, and robust loss functions. Label correction methods aim to identify and fix mislabeled data, often by modeling noise or using auxiliary networks to infer the true labels~\cite{xiao2015learning, zheng2021meta,yu2024noise}. Also, combining label correction with loss reweighting has been shown to improve performance under high noise rates~\cite{li2024sure, englesson2024robust}. Sample selection techniques focus on choosing reliable data points to reduce the influence of noisy data, using strategies like curriculum learning to progressively emphasize cleaner samples or contrastive learning to highlight informative examples~\cite{jiang2018mentornet, wei2021jo,song2019selfie,wu2024topological}.  Refined training methodologies adapt the learning process to tolerate label noise; for example, some approaches train multiple models together to exchange high-confidence predictions (co-teaching), while others treat noisy data as unlabeled and leverage semi-supervised learning techniques~\cite{han2018co,li2020dividemix,tanaka2018joint}. However, many of the above approaches rely on auxiliary networks, multiple model training, or assumptions about the noise distribution, making them computationally expensive and harder to implement. Additionally, they often require careful hyperparameter tuning and clean validation data, which limits their practicality in real-world settings. By contrast, robust loss functions embed noise tolerance directly into the training objective, offering a simpler yet powerful alternative. They are generally model-agnostic and easy to integrate into existing pipelines, making them particularly suitable for scalable and end-to-end learning.

The theoretical robustness of loss functions in the presence of label noise has been established in prior work~\cite{ghosh2017robust}, which shows that the widely used Cross-Entropy (CE) loss is not noise-tolerant, while the Mean Absolute Error (MAE) loss satisfies robustness conditions. However, despite their theoretical robustness, losses like MAE tend to converge slowly, underfit the data, or even fail to converge altogether \cite{ma2020normalized}. This highlights that robustness is only one side of the trade-off, the other being convergence properties.

To overcome the convergence problem of robust losses, researchers have developed combined loss functions that integrate the strengths of robust losses (e.g., MAE) with those of fast-converging losses (e.g., CE) \cite{feng2021can,mao2023cross}. In \cite{zhang2018generalized}, a Box-Cox transformation has been used to combine MAE and CE, employing a hyperparameter to shift the loss behavior from CE-like to MAE-like. Later, MAE-like and CE-like loss functions were categorized as “passive” and “active” losses, respectively, based on their optimization behaviors \cite{ma2020normalized}. Active loss functions explicitly maximize the classifier’s output probability at the true label position, whereas passive losses additionally minimize the probability of at least one other class position. This categorization led to the state-of-the-art Active-Passive Loss (APL) framework, which is defined as the weighted summation of an active and a passive loss. APL addresses the underfitting issues of theoretically robust loss functions. Many loss functions designed for label noise problems naturally fit within the APL framework. For example, the Symmetric Cross-Entropy (SCE) loss \cite{wang2019symmetric} can be interpreted in this context. Inspired by symmetric Kullback–Leibler divergence, SCE employs an information-theoretic perspective by introducing a noise-tolerant term, Reverse Cross-Entropy (RCE). By combining CE and RCE with appropriate coefficients, SCE effectively enhances performance on datasets containing label noise. In \cite{zhou2023asymmetric}, a novel category of asymmetric loss functions is proposed to enhance noise tolerance in both classification and regression tasks. Another study \cite{englesson2021generalized} proposed using the Jensen–Shannon divergence as a noise-robust loss function, demonstrating its ability to interpolate between CE and MAE through a controllable mixing parameter. Another approach \cite{ye2023active} introduced active-negative loss functions, which generate robust passive losses from existing active losses. However, these methods share a crucial drawback: they require hyperparameter tuning, one for GCE and two for APL schemes, to handle label noise effectively, typically via trial and error for each dataset. Poor tuning of these hyperparameters can severely degrade performance or even cause convergence failures.
\subsection{Main Contributions}
 In this paper, we propose Fractional Classification Loss (FCL), which leverages the APL framework while eliminating the need for manual coefficient tuning. Specifically, we introduce a new active loss, Fractional Cross-Entropy (FCE), obtained by applying the fractional derivative to the CE loss with respect to the negative log-likelihood, while the passive loss is the MAE. This formulation reduces the APL coefficients to a single robustness hyperparameter, the fractional derivative order \(\mu\), which smoothly transitions FCL between MAE-like robustness and CE-like convergence. Increasing \(\mu\) decreases the log penalty and gradient magnitude on noisy labels, enhancing robustness at the expense of convergence (MAE-like behavior). Conversely, decreasing \(\mu\) results in faster convergence but less robustness against label noise (CE-like behavior). Despite reducing the APL coefficients to one interpretable parameter, determining the optimal \(\mu\) before training remains challenging, especially for large datasets where label noise information is typically unavailable.

To address this challenge, we propose transforming \(\mu\) into a learnable parameter and integrating it into gradient-based optimization. We reveal that the unique property of FCL presents a crucial trade-off that enables the stable learning of \(\mu\): lower log penalties on difficult or mislabeled examples improve robustness but impose higher penalties on easy or clean data, reducing model confidence in them. This trade-off enables stable learning of \(\mu\) during DNN training based on the proportion of mislabeled data, tuning \(\mu\) higher in noisier datasets and lower in cleaner ones. This adaptive mechanism maintains an optimal balance between robustness and convergence speed, allowing FCL to adjust to any dataset without manual hyperparameter tuning. 

To assess the effectiveness of FCL, we conduct ablation studies and first demonstrate that mistuned hyperparameters degrade performance in existing robust losses, an issue avoided by FCL’s adaptive learning of \(\mu\). Next, we evaluate performance across different initializations of \(\mu\) and compare models with fixed versus adaptive \(\mu\), in addition to assessing the time complexities of existing losses and FCL. Finally, experiments on MNIST, CIFAR-10, and CIFAR-100 with both symmetric and asymmetric label noise confirm that FCL consistently achieves state-of-the-art performance while automatically tuning \(\mu\) for robustness.

\subsection{Organization of the Study}
The remainder of the paper is organized as follows. In Section \ref{sec: Classification with Noisy Labels}, we introduce the classification problem under noisy labels and review existing loss functions. Section \ref{sec:fcl} details the derivation and properties of the proposed FCL. In Section \ref{sec: Learning Robust and Adaptive FCL}, we describe our approach for learning the fractional derivative order $\mu$, which enables FCL to dynamically adjust its robustness based on the noise level in the data. Section \ref{sec:experiments} presents ablation studies and comprehensive experiments on benchmark datasets under both symmetric and asymmetric label noise, comparing FCL with state-of-the-art loss functions. Finally, Section \ref{sec: Conclusion and Future Work} concludes the paper by summarizing our key contributions and discussing directions for future research.
\section{Classification with Noisy Labels}\label{sec: Classification with Noisy Labels}
\subsection{Problem Description}
We consider a standard \(K\)-class classification problem defined over a feature space \(\mathcal{X} \subset \mathbb{R}^d\) denoting a sample in the $d$-dimensional input space and a label set \(\mathcal{Y} = [k] = \{1,\ldots,K\}\). A clean dataset \(\mathcal{S} = \{(x_n, y_n)\}_{n=1}^N\) is drawn i.i.d. from an unknown distribution \(\mathcal{D}\) over \(\mathcal{X} \times \mathcal{Y}\). For a given sample \(x\), let \(\mathbf{q}(k|x)\) denote the true probability distribution over labels, satisfying \(\sum_{k=1}^{K} \mathbf{q}(k|x) = 1\). In the clean setting, each sample \(x\) has a unique correct label, meaning \(\mathbf{q}(y_n|x_n) = 1\) and \(\mathbf{q}(k\neq y_n|x_n) = 0\); that is, \(\mathbf{q}(k|x)\) is simply the one-hot encoding of the label.

We define a classifier \(h(x) = \arg\max_{i} f_i(x)\), where \(f: \mathcal{X} \rightarrow \mathcal{C}\) maps the feature space to the label space. In our work, we consider \(f\) as a DNN ending with a softmax output layer. For each sample \(x\), the network produces a probability distribution over classes:
\begin{equation}
\boldsymbol{p}(k|x) = \frac{\exp\bigl(f_k(x)\bigr)}{\sum_{j=1}^K \exp\bigl(f_j(x)\bigr)},
\end{equation}
with \(\sum_{k=1}^{K} \boldsymbol{p}(k|x) = 1\). Throughout, we refer to \(f\) as the classifier. Training a classifier \(f\) involves finding an optimal classifier \(f^*\) that minimizes the empirical risk,
\begin{equation}
R_{L}(f) = \mathbb{E}_{\mathcal{D}}\left[L\bigl(f(x), y\bigr)\right] = \frac{1}{N} \sum_{n=1}^N L\bigl(f(x_n), y_n\bigr).
\end{equation}
where \(L\colon \mathcal{C} \times \mathcal{Y} \to \mathbb{R}^{+}\) is a loss function, and \(L(f(x), k)\) denotes the loss of \(f(x)\) with respect to label \(k\).

In many practical scenarios, the training set is corrupted by label noise, and only a noisy dataset \(\mathcal{S}_{\eta} = \{(x_n, \hat{y}_n)\}_{n=1}^N\) is available, with samples drawn i.i.d. from an unknown distribution \(\mathcal{D}_{\eta}\), and $\hat{y}_n$ are the noisy labels. A common model for label noise assumes that, given the true label \(y\), the corruption process is conditionally independent of the input \(x\). Specifically, the true label \(y\) is flipped to a different label \(j \neq y\) with probability \(\eta_{yj}\), resulting in an overall noise rate \(\eta_y = \sum_{j\neq y}\eta_{yj}\). The noise is said to be \emph{symmetric} if \(\eta_{ij} = \frac{\eta_i}{K-1}\) for all \(j \neq y\) with \(\eta_i = \eta\), where \(\eta \in [0,1]\). For \emph{asymmetric} noise, the corruption probabilities \(\eta_{ij}\) depend on both the true label \(i\) and the corrupted label \(j\). Under noise, the corresponding risk becomes
\begin{equation}
R_{L}^{\eta}(f) = \mathbb{E}_{\mathcal{D}_{\eta}}\left[L\bigl(f(x), \hat{y}\bigr)\right].
\end{equation}

A loss \(L\) is called robust if its global minimizer on clean data also minimizes \(R_{L}^{\eta}(f)\) when labels are noisy. Several works propose such robust loss functions and demonstrate how a loss function can be robust to label noise \cite{ghosh2017robust}. However, all those works have observed slow convergence of such robust loss functions leading to underfitting and result in even worse performance than nonrobust loss functions \cite{ma2020normalized}. Therefore, there is a trade-off between robustness and convergence for loss function design.

\subsection{Existing Classification Loss Functions} 
The most commonly used classification loss CE is defined for a single training sample \(x\) as:
\begin{equation}
L_{CE}(f(x), y)
\;=\;
-\sum_{k=1}^K
\boldsymbol{q}(k|x)\,\log\,\boldsymbol{p}(k|x),
\end{equation}
Its fast convergence is attributed to the high log penalty imposed on difficult samples (i.e., those with low \(\boldsymbol{p}(k|x)\) values for true class k) \cite{ma2020normalized}. However, despite its popularity, CE is known to be non-robust under label noise \cite{ghosh2017robust}. 

An alternative is the MAE loss, given by
\begin{equation}
L_{MAE}(f(x), y)
\;=\;
\sum_{k=1}^K
\bigl|\,\boldsymbol{p}(k|x) - \boldsymbol{q}(k|x)\bigr|.
\end{equation}
MAE has been shown to be provably robust to label noise \cite{ghosh2017robust}, but it suffers from slow convergence in practice due to its relatively weak penalization of difficult samples \cite{ma2020normalized}.

To combine the benefits of CE’s implicit weighting of difficult samples with MAE’s robustness, the Generalized Cross-Entropy (GCE) loss was introduced \cite{zhang2018generalized}. GCE is defined as the negative Box-Cox transformation bridging CE and MAE:
\begin{equation}
L_{GCE}(f(x), y)
\;=\;
\sum_{k=1}^K \frac{1 - \boldsymbol{p}(k|x)^{\,q}}{q},
\end{equation}
with \(q \in (0,1]\). As \(q \to 0\), GCE recovers the standard CE loss; for \(q=1\), it behaves similarly to MAE, thereby providing a tunable balance between convergence speed and robustness.

Active-passive loss functions have been proposed in \cite{ma2020normalized}. Active losses explicitly maximize the network’s output probability at the class position specified by the label \(y\) (e.g., CE loss), while passive losses explicitly minimize the probability at one or more of the other class positions (e.g., MAE loss). These are combined as \cite{ma2020normalized}:
\begin{equation}
\label{eq:active-passive}
L_{\mathrm{APL}}
\;=\;
\alpha\,L_{\mathrm{Active}}
\;+\;
\beta\,L_{\mathrm{Passive}},
\end{equation}
where \(\alpha \in \mathbb{R}\) and \(\beta \in \mathbb{R}\) are coefficients that balance the contributions of the active and passive losses.

Another partially robust alternative is the SCE loss \cite{wang2019symmetric}, which combines CE with RCE:
\begin{equation}
\label{eq: sce}
L_{SCE} = \alpha L_{CE} + \beta L_{RCE}.
\end{equation}
The RCE loss reverses the roles of \(p\) and \(q\) in the logarithmic term and is defined as
\begin{equation}
L_\mathrm{RCE}(x)
\;=\;
-\sum_{k=1}^K
\boldsymbol{p}(k|x)\,\log\,\boldsymbol{q}(k|x).
\end{equation}

Overall, these approaches illustrate a central theme in robust classification: balancing a fast-converging, noise-sensitive component (e.g., CE) with a noise-robust, slow-converging component (e.g., MAE or RCE) to enhance performance under label noise. However, they all require tuning additional hyperparameters, whether the \(q\) parameter in GCE or the \(\alpha\) and \(\beta\) coefficients in active–passive losses (or SCE, which can be seen as a variant within this framework), typically through trial and error, as the label noise level is unknown apriori.

\section{Fractional Classification Loss}\label{sec:fcl}

In this section, we present the derivation and properties of FCL, an adaptive and robust loss function parameterized by the fractional derivative order \(\mu\). This parameter controls the balance between robustness to label noise and convergence speed. We denote the basic component of the loss \(L(f(x), y)\) as \(\ell(f(x), k)\), so that $L(f(x), y)=\sum_{k=1}^K \ell(f(x), k)$. We define FCL within the active-passive loss framework (see \eqref{eq:active-passive}) but without the coefficients \(\alpha\) and \(\beta\):
\begin{equation}
\label{eq: fractional classification loss}
\ell_{FCL}^{\mu}(f(x), y) \;:=\; \ell_{FCE}^{\mu}(f(x),y)\;+\;\ell_{MAE}(f(x),y).
\end{equation}
Here, the passive loss function is the MAE, denoted by $\ell_{MAE}$. For the active loss, we propose $\ell_{FCE}^{\mu}$, defined as the fractional derivative of the CE loss with respect to the negative log-likelihood, $-\log \boldsymbol{p}(k|x)$. Specifically, let $u\bigl(\boldsymbol{p}(k|x)\bigr) = -\log\bigl(\boldsymbol{p}(k|x)\bigr)$, then we can define the CE loss as a power function, i.e., $\ell_{CE}(u) = u$. By applying the fractional derivative for power functions (see Appendix for details), we define the FCE as:
\begin{equation}
\label{eq: fractional cross entropy}
\ell_{FCE}^{\mu}(f(x), y) 
\;:=\;
\frac{\partial^\mu\,\ell_{CE}(f(x),y)}{\partial \left[-\log \boldsymbol{p}(k|x)\right]^\mu}
\;=\;
\frac{\left[-\log \boldsymbol{p}(k|x)\right]^{1-\mu}}{\Gamma(2-\mu)}.
\end{equation}
By taking the fractional derivative with respect to \(-\log \boldsymbol{p}(k|x)\), the parameter \(\mu\) determines the extent to which the logarithmic penalty is applied for values of \(\boldsymbol{p}(k|x)\).

We examine two special cases to illustrate the effect of \(\mu\):
\begin{itemize}
    \item \(\mu = 0\). In this case, \(\ell_{FCE}^{\mu}\) reduces to the CE loss, and FCL becomes:
    \begin{equation}
    \label{eq: fcl when mu 0}
    \ell_{FCL}^{0}(f(x),y)
    \;=\;
    \ell_{CE}(f(x),y)\;+\;\ell_{MAE}(f(x),y).
    \end{equation}
    This formulation applies the full logarithmic penalty to low values of \(\boldsymbol{p}(k|x)\), reflecting the implicit weighting of difficult examples by CE, which accelerates convergence but increases sensitivity to label noise.
    
    \item \(\mu = 1\). Here, the FCE term becomes a constant (equal to 1), so FCL reduces to a shifted MAE:
    \begin{equation}
    \label{eq: fcl when mu 1}
    \ell_{FCL}^{1}(f(x),y)
    \;=\;
    \ell_{MAE}(f(x),y)\;+\;1.
    \end{equation}
    This yields a loss that applies uniform penalties across all samples, characteristic of MAE, thereby increasing robustness but resulting in slower convergence due to the diminished penalty of difficult examples.
\end{itemize}
\begin{figure}[t]
  \centering
  \subfloat[]{\includegraphics[scale=0.49]{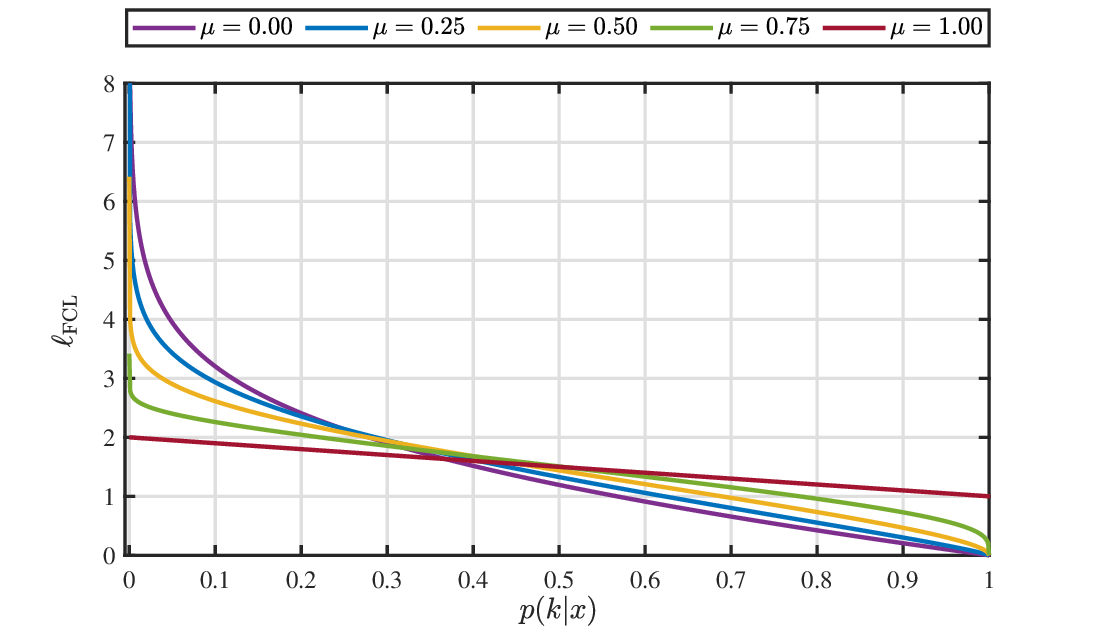}
  \label{fig:FCL loss plot}}
  \\
  \vspace{-0.35cm}
  \subfloat[]{\includegraphics[scale=0.49]{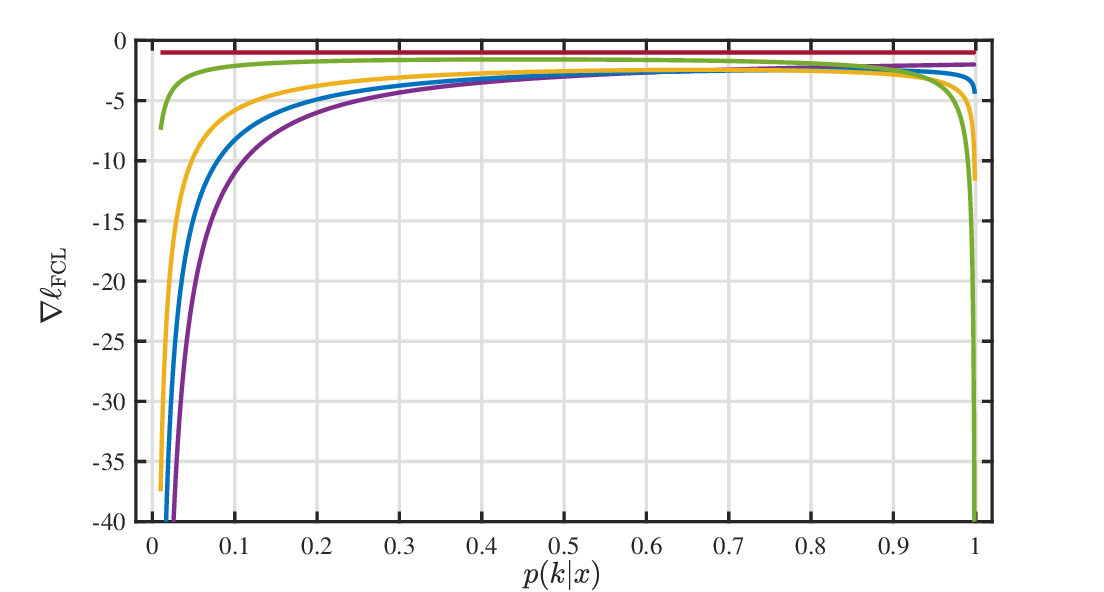}
  \label{fig:grad of FCL}}
  \caption{(a) The FCL curve under different values of \(\mu\), and (b) its gradient.}
  \label{fig: FCL combined}
\end{figure}
 Fig.~\ref{fig:FCL loss plot} illustrates the behavior of FCL for the true class probability \(\boldsymbol{p}(k|x)\) as \(\mu\) varies. When \(\mu\) is small, a heavy logarithmic penalty is imposed on low values of \(\boldsymbol{p}(k|x)\) (i.e., for difficult samples or those affected by label noise). As \(\mu\) approaches 1, this penalty decreases, making the loss function more insensitive to noisy samples, though at the expense of convergence speed. Thus, \(\mu \in [0,1]\) continuously interpolates between a CE–like loss and an MAE–like loss.

\subsection{Gradient of the FCL}

In this section, we analyze \(\ell_{FCL}\) from the perspective of gradient-based optimization. We first consider the gradient of the active component, \(\ell_{FCE}^{\mu}\). From \eqref{eq: fractional cross entropy}, since \(\Gamma(2-\mu)\) does not depend on \(\boldsymbol{p}(k|x)\), we treat it as a constant and define $u\bigl(\boldsymbol{p}(k|x)\bigr) = -\log\bigl(\boldsymbol{p}(k|x)\bigr)$. Applying the power rule, we obtain:
\begin{equation}
\frac{\partial}{\partial\,\boldsymbol{p}(k|x)}\Bigl[u\bigl(\boldsymbol{p}(k|x)\bigr)^{1-\mu}\Bigr]
= \frac{(\mu-1) \Bigl[-\log\bigl(\boldsymbol{p}(k|x)\bigr)\Bigr]^{-\mu}}{\boldsymbol{p}(k|x)} .
\end{equation}
Thus, the derivative of \(\ell_{FCE}^{\mu}\) with respect to \(\boldsymbol{p}(k|x)\) is:
\begin{equation}
    \nabla \ell_{FCE}^{\mu}
    \;=\;
    \frac{1}{\Gamma(2-\mu)}\left[-\,\frac{1-\mu}{\boldsymbol{p}(k|x)} \Bigl[-\log\bigl(\boldsymbol{p}(k|x)\bigr)\Bigr]^{-\mu}\right].
\end{equation}
For the MAE component, the derivative with respect to \(\boldsymbol{p}(k|x)\) (for the target class) is simply constant \(-1\). Combining these results yields the total gradient of FCL:
\begin{equation}
\label{eq: gradient w.r.t input}
    \nabla \ell_{FCL}
    \;=\;
    -\,\frac{1-\mu}{\Gamma(2-\mu)} \, \frac{\Bigl[-\log\bigl(\boldsymbol{p}(k|x)\bigr)\Bigr]^{-\mu}}{\boldsymbol{p}(k|x)}
    \;-\; 1.
\end{equation}

Figure~\ref{fig:grad of FCL} illustrates how gradient \eqref{eq: gradient w.r.t input} changes with \(\mu\). When \(\mu = 0\), difficult or mislabeled samples (i.e., those with small \(\boldsymbol{p}(k|x)\)) receive larger gradient magnitudes, which speeds up convergence but increases the risk of overfitting to noisy labeled data. As \(\mu\) increases toward 1, the gradient on these hard examples reduces, shifting FCL toward MAE’s uniform treatment of all data samples and thereby enhancing robustness to noisy labels, but with a slower convergence.

\subsection{Linking Fractional Derivative Order to APL Coefficients}

Here, we show the relationship between the fractional derivative order \(\mu\) and the APL coefficients $\alpha$ and $\beta$. Specifically, we demonstrate the robustness effect arising from the  \(\beta\) coefficient in SCE and link it to $\mu$. Recall that SCE~\eqref{eq: sce} is an APL in which the coefficients \(\alpha\) and \(\beta\) control the relative weighting of the CE (active loss) and RCE (passive loss) terms \cite{wang2019symmetric}. In SCE, the RCE term is a scaled version of MAE \cite{ye2023active}:
\begin{equation}
    \ell_{RCE}(f(x),y) = \frac{-A}{2}\,\ell_{MAE}(f(x),y),
\end{equation}
\begin{equation}
\label{eq: SCE with MAE}
    \ell_{SCE}(f(x),y) 
    \;=\;
    \alpha\,\ell_{CE}(f(x),y) \;+\; \frac{-A\cdot\beta}{2}\,\ell_{MAE}(f(x),y),
\end{equation}
where \(A<0\) is a constant defined to prevent \(\log 0\) (i.e., \(\log 0 = A\)). Increasing \(\alpha\) places more emphasis on the active component CE, while increasing \(\beta\) places more weight on the MAE term, thereby making the loss more robust to label noise but converging more slowly.

Dividing \eqref{eq: SCE with MAE} by \(\tfrac{-A\cdot\beta}{2}\) yields
\begin{equation}
\label{eq: sce with mae form normalized}
    \frac{\ell_{SCE}(f(x),y)}{\frac{-A\cdot\beta}{2}} 
    \;=\; 
    \frac{-2\,\alpha}{A\,\beta}\,\ell_{CE}(f(x),y) 
    \;+\; 
    \ell_{MAE}(f(x),y).
\end{equation}

Next, rewriting \(\ell_{FCE}^{\mu}\) in \eqref{eq: fractional cross entropy} by isolating the cross-entropy term \(-\log \boldsymbol{p}(k|x)\) gives
\begin{equation}
\label{eq: coeff seperated fce}
    \ell_{FCE}^{\mu}(f(x),y) 
    \;=\; 
    \frac{1}{\Gamma(2-\mu)\,\bigl[-\log \boldsymbol{p}(k|x)\bigr]^{\mu}}
    \,\bigl[-\log \boldsymbol{p}(k|x)\bigr].
\end{equation}
Since our FCL formulation in \eqref{eq: fractional classification loss} effectively sets the MAE coefficient to 1, we can equate the CE coefficient in \eqref{eq: sce with mae form normalized} to the CE coefficient in \eqref{eq: coeff seperated fce}. This matching leads to the relation between \(\mu\) and \(\beta\):
\begin{equation}
\label{eq: equality mu and beta}
    \frac{-2\alpha}{A\cdot\beta} 
    \;=\; 
    \frac{1}{\Gamma(2-\mu)\,\bigl[-\log \boldsymbol{p}(k|x)\bigr]^{\mu}}.
\end{equation}
For \(\alpha=1\), we obtain
\begin{equation}
    \beta 
    \;=\; 
    \frac{2\,\Gamma(2-\mu)\,\bigl[-\log \boldsymbol{p}(k|x)\bigr]^{\mu}}{A}.
\end{equation}

\begin{figure}[t!]
    \centering
    \includegraphics[width=0.55\linewidth]{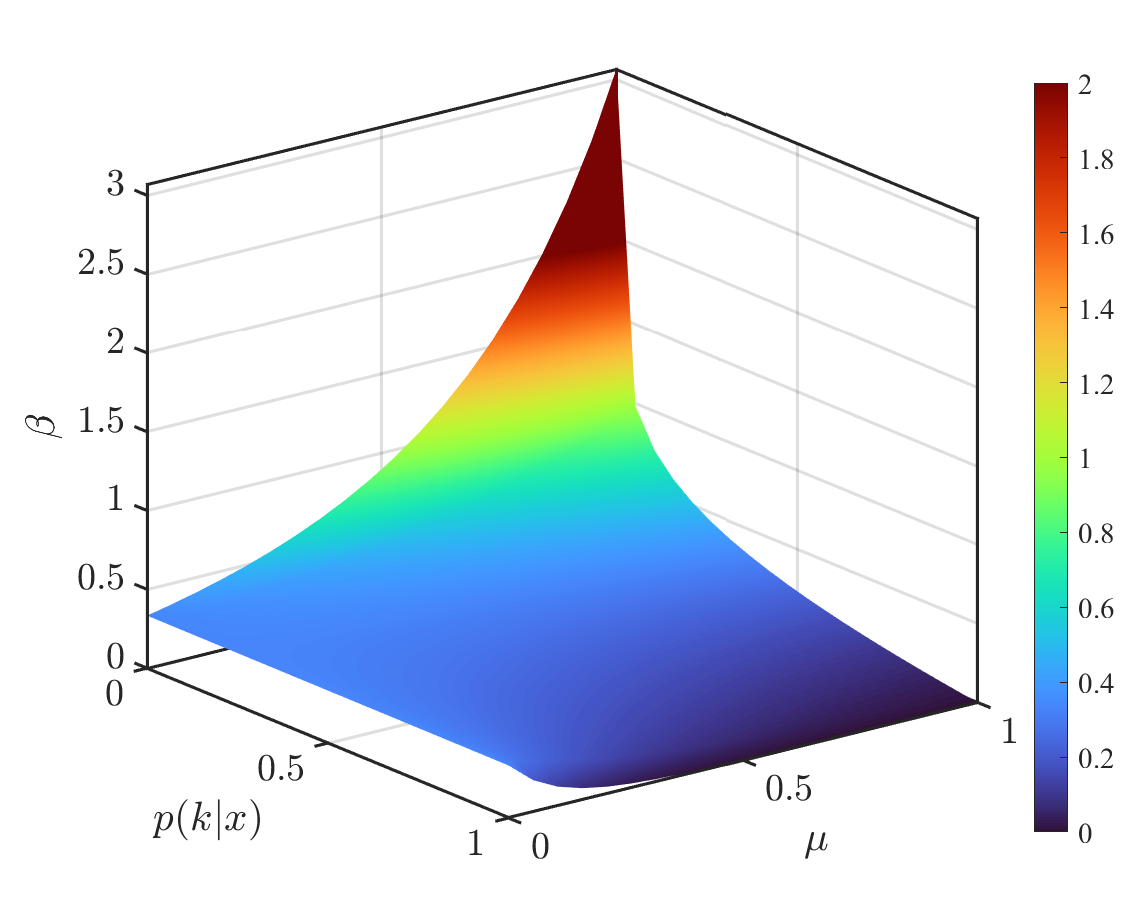}
    \caption{A 3D view of \(\beta\) as a function of \(\mu\) and the softmax probability \(\boldsymbol{p}(k|x)\).}
    \label{fig:enter-label}
\end{figure}

Fig.~\ref{fig:enter-label} plots \(\beta\) as a function of \(\mu\) and \(\boldsymbol{p}(k|x)\) (using \(A=-6\) as in \cite{wang2019symmetric}). For small \(\boldsymbol{p}(k|x)\) (i.e., for difficult or noisy samples), increasing \(\mu\) corresponds to raising \(\beta\), thereby making FCL more robust to label noise. Consequently, FCL requires only a single hyperparameter, \(\mu\), instead of two (\(\alpha\) and \(\beta\)) to tune. In the next section, we show that FCL possesses the necessary properties to learn \(\mu\) during training, eliminating the need for any manual hyperparameter tuning.
\section{Learning Robust and Adaptive FCL}\label{sec: Learning Robust and Adaptive FCL}
We now extend the proposed FCL by treating the fractional derivative order \(\mu\) as a learnable parameter, thereby transforming \(\ell_{\mathrm{FCL}}(f(x), y)\) into \(\ell_{\mathrm{FCL}}(f(x), y, \mu)\). This enables the empirical risk minimization process to automatically adjust \(\mu\) during training to optimize the trade-off between robustness and convergence:
\begin{equation}
R_{L}(f, \mu) = \mathbb{E}_{\mathcal{D}}\left[L\bigl(f(x), y, \mu\bigr)\right] = \frac{1}{N} \sum_{n=1}^N L\bigl(f(x_n), y_n, \mu\bigr).
\end{equation}

In traditional APLs, the hyperparameters \(\alpha\) and \(\beta\) are manually tuned. Attempts to learn these parameters via gradient-based optimization tend to drive them toward \(-\infty\), thereby minimizing the loss without meaningful model learning. In contrast, FCL exhibits an inherent trade-off that allows stable learning of its hyperparameter, fractional derivative order \(\mu\).

This trade-off is illustrated in Fig.~\ref{fig:FCL loss plot}. As \(\mu\) increases, the logarithmic penalty applied to difficult or noisy samples (i.e., those with small values of \(\boldsymbol{p}(k|x)\)) decreases, enhancing robustness to label noise. However, simultaneously, a higher penalty is imposed on easy-to-classify samples (i.e., with large values of \(\boldsymbol{p}(k|x)\)). A similar behavior is observed in the gradients in Fig.~\ref{fig:grad of FCL}: with increasing \(\mu\), the gradients for samples with small \(\boldsymbol{p}(k|x)\) decrease, while for samples with large \(\boldsymbol{p}(k|x)\) increase. This creates a critical trade-off, the model must be sufficiently robust to handle noisy labels without sacrificing its performance on easy-to-classify examples.

During DNN training, the fractional derivative order $\mu$ is learned based on the amount of label noise in the dataset. Difficult or noisy samples, typically misclassified by the model, tend to have low \(\boldsymbol{p}(k|x)\) and appear on the left side of the loss curve (Fig.~\ref{fig:FCL loss plot}), whereas easy or correctly labeled examples are on the right (where $k$ is the true class label). This trade-off is critical when learning $\mu$ during training. In contrast, if the $\alpha$ and $\beta$ parameters in APLs are treated as learnable, gradient descent tends to push them toward $-\infty$ in an attempt to minimize the loss indefinitely, effectively collapsing the training and preventing meaningful learning.

As shown in Section \ref{sec:fcl}, increasing \(\mu\) flattens the left side of the loss curve (reducing the penalty on noisy data) and steepens the right side (increasing the penalty on easy-to-classify examples). Thus, in datasets with many noisy labels, the \(\mu\) will converge to higher values to enhance robustness, while in datasets with fewer noisy labels, \(\mu\) will converge to lower values to avoid excessive penalization of easy samples. In this way, FCL adapts its robustness in accordance with the noise level in the dataset, avoiding the pitfalls of excessive robustness such as slow convergence and underfitting.
\begin{algorithm}[t]
\caption{Training with Fractional Classification Loss}
\label{alg:fractional-classification}
\begin{algorithmic}[1]
\REQUIRE Training set \(\mathcal{D}_{\eta}\) (inputs \(\mathbf{x}\), noisy labels \(\hat{\mathbf{y}}\)), initial network parameters \(\theta\), initial fractional parameter \(\mu(0)\), learning rates \(\lambda_{\theta}\) and \(\lambda_{\mu}\), number of epochs \(E\), mini-batch size \(B\).
\STATE Randomly initialize \(\theta\).
\STATE Set \(\mu \leftarrow \mu(0)\).
\FOR{\(e = 1\) \TO \(E\)}
    \STATE Set \(\mathrm{accGrad} \leftarrow 0\).
    \FOR{each mini-batch \(\mathcal{B} \subset \mathcal{D}_{\eta}\) of size \(B\)}
        \STATE Compute \(\tilde{\mathbf{y}} = f_{\theta}(\mathbf{x})\).
        \STATE Compute loss \(\ell \leftarrow \ell_{\mathrm{FCL}}(\tilde{\mathbf{y}}, y; \mu)\).
        \STATE Backpropagate to obtain \(\nabla_{\theta}\ell\) and \(\nabla_{\mu}\ell\).
        \STATE Update network parameters: 
        \[
          \theta \leftarrow \theta - \lambda_{\theta}\,\nabla_{\theta}\ell.
        \]
        \STATE Accumulate gradient for \(\mu\): 
        \[
          \mathrm{accGrad} \leftarrow \mathrm{accGrad} + \nabla_{\mu}\ell.
        \]
    \ENDFOR
    \STATE Update fractional parameter:
    \[
      \mu \leftarrow \mu - \lambda_{\mu}\,\left(\frac{\mathrm{accGrad}}{B}\right).
    \]
\ENDFOR
\end{algorithmic}
\end{algorithm}
We now derive the gradient of the FCE loss with respect to the \(\mu\). First, we define \(\ell_{\mathrm{FCE}}(\mu) = \frac{r(\mu)}{g(\mu)}\) where
\begin{equation}
\label{eq:fce_f}
r(\mu) 
\;=\;
\bigl[-\log \boldsymbol{p}(k|x)\bigr]^{\,1-\mu}, \nonumber
\end{equation}
\begin{equation}
\label{eq:fce_g}
g(\mu)
\;=\;
\Gamma\bigl(2-\mu\bigr).
\end{equation}
We can write the $r'(\mu)$ as:
\begin{IEEEeqnarray}{rCl}
\label{eq:fce_fprime}
r'(\mu) &=& \frac{\partial}{\partial \mu}\,
\bigl[-\log \boldsymbol{p}(k|x)\bigr]^{\,1-\mu} \nonumber\\[1mm]
        &=& -\log\!\bigl[-\log \boldsymbol{p}(k|x)\bigr] \,\bigl[-\log \boldsymbol{p}(k|x)\bigr]^{\,1-\mu}.
\end{IEEEeqnarray}
For \(g'(\mu)\), we use the relation \(\frac{d}{dz}\Gamma(z) = \Gamma(z)\,\Psi(z)\), where \(\Psi(\cdot)\) is the digamma function. Thus,
\begin{IEEEeqnarray}{rCl}
\label{eq:fce_gprime}
g'(\mu) = \frac{\partial}{\partial \mu}\,\Gamma\bigl(2-\mu\bigr)-\,\Gamma\bigl(2-\mu\bigr)\,\Psi\bigl(2-\mu\bigr)
\end{IEEEeqnarray}
By applying the quotient rule for derivative, and factoring out \(\bigl[-\log \boldsymbol{p}(k|x)\bigr]^{\,1-\mu}\,\Gamma\bigl(2-\mu\bigr)\), we obtain
\begin{IEEEeqnarray}{rCl}
\label{eq: grad w.r.t mu}
&&\frac{\partial}{\partial\mu}\,\ell_{\mathrm{FCE}}(\mu)
 =  \nonumber \\ 
&& \frac{\bigl[-\log \boldsymbol{p}(k|x)\bigr]^{\,1-\mu}}{\Gamma(2-\mu)} \Bigl[\Psi\bigl(2-\mu\bigr) - \log\!\bigl(-\log \boldsymbol{p}(k|x)\bigr)\Bigr].
\end{IEEEeqnarray}
This derivative can be computed via backpropagation using standard automatic differentiation workflows, which is particularly useful in scenarios where explicit gradient definitions are required.

Algorithm~\ref{alg:fractional-classification}\footnote[1]{MATLAB implementation. [Online]. Available: GitHub repo will be shared upon the acceptance of the paper} outlines the complete training procedure. For stable learning of \(\mu\), we update it once per training epoch rather than at every mini-batch. During each epoch, we accumulate the gradient with respect to \(\mu\) and then update \(\mu\) using a gradient-based optimizer (e.g., Adam or SGD) with a relatively large learning rate compared to that for the model parameters to ensure stable convergence. Additionally,  we keep $\mu$ fixed for the first 5 epochs before updating it.

\section{Experiments}
\label{sec:experiments}
In this section, we evaluate the effectiveness of our proposed FCL on benchmark datasets MNIST, CIFAR-10, and CIFAR-100 under both symmetric and asymmetric label noise. We compare FCL with several state-of-the-art robust and non-robust loss functions. Details of the experimental setup, including network architectures, training procedures, and parameter settings, are provided. Ablation studies are conducted to demonstrate the empirical properties of FCL. All experiments are repeated three times using random seeds for all loss functions to ensure reproducibility. All experiments were conducted within MATLAB\textsuperscript{\textregistered} via MATLAB's Deep Learning Toolbox\textsuperscript{\texttrademark}

\subsection{Ablation Studies}
To investigate the properties of the proposed FCL, we performed a series of ablation studies on the CIFAR-10 dataset. For these experiments, a validation set comprising 20\% of the training data was randomly sampled, and symmetric label noise was introduced at a rate of 0.6. This high noise rate of $60\%$ serves as a stress test scenario, which is well-suited for ablation studies, as conventional loss functions often struggle by either overfitting to noisy labels or failing to learn meaningful patterns.

\subsubsection{Impact of Hyperparameter Tuning on Existing Robust Losses}
\begin{figure}[b!]
    \centering
    \includegraphics[width=0.55\linewidth]{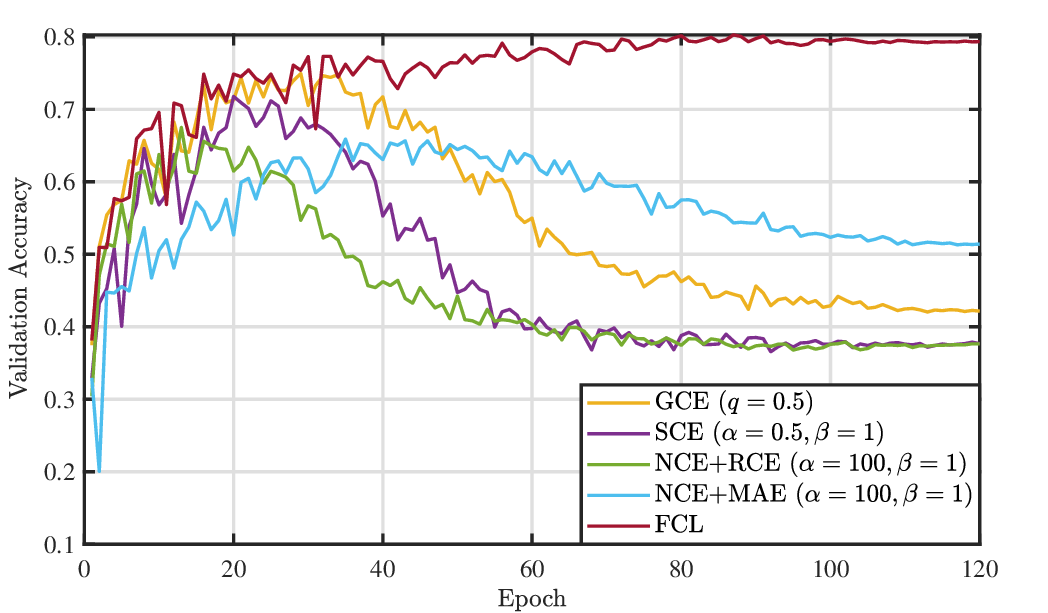}
    \caption{Validation accuracies of baseline robust loss functions (GCE, SCE, NCE+MAE, and NCE+RCE) with mistuned hyperparameters versus FCL.}
    \label{fig:mistune hyperparam}
\end{figure}

We first evaluated existing robust loss functions, including GCE, SCE, Normalized Cross-Entropy (NCE)+MAE, and NCE+RCE, under mistuned hyperparameter settings. Similar evaluations are presented in the original works. For our experiments, we adopt the mistuned hyperparameter values as reported by the studies \cite{ma2020normalized, zhang2018generalized,wang2019symmetric}. As shown in Fig.~\ref{fig:mistune hyperparam}, improper selection of the robustness hyperparameters \(q\), \(\alpha\), and \(\beta\) leads to a substantial decrease in performance. To achieve good accuracy with baseline methods, an extensive grid search is typically required to find optimal $\alpha$ and $\beta$ hyperparameters. In contrast, FCL consistently delivers superior performance without the risk of mistuning, as it automatically adapts its robustness during training.
\subsubsection{Effect of Initial Values for \(\mu\)}
\begin{figure}[t]
  \centering
  \subfloat[]{\includegraphics[scale=0.49]{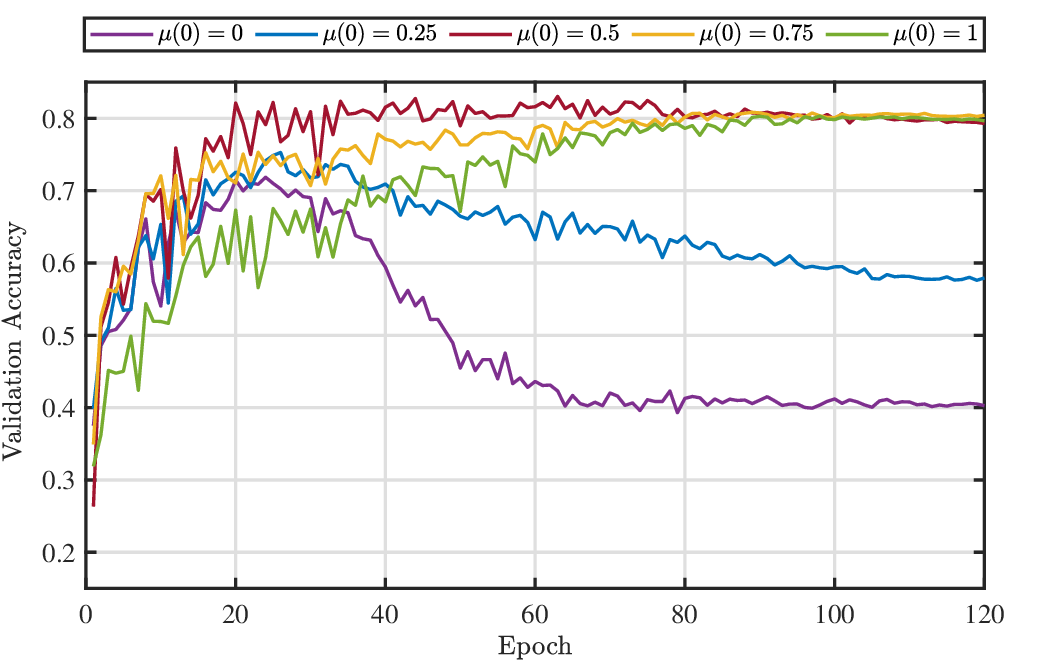}
  \label{fig:initial mu accuracy}}
  \\
  \vspace{-0.35cm}
  \subfloat[]{\includegraphics[scale=0.49]{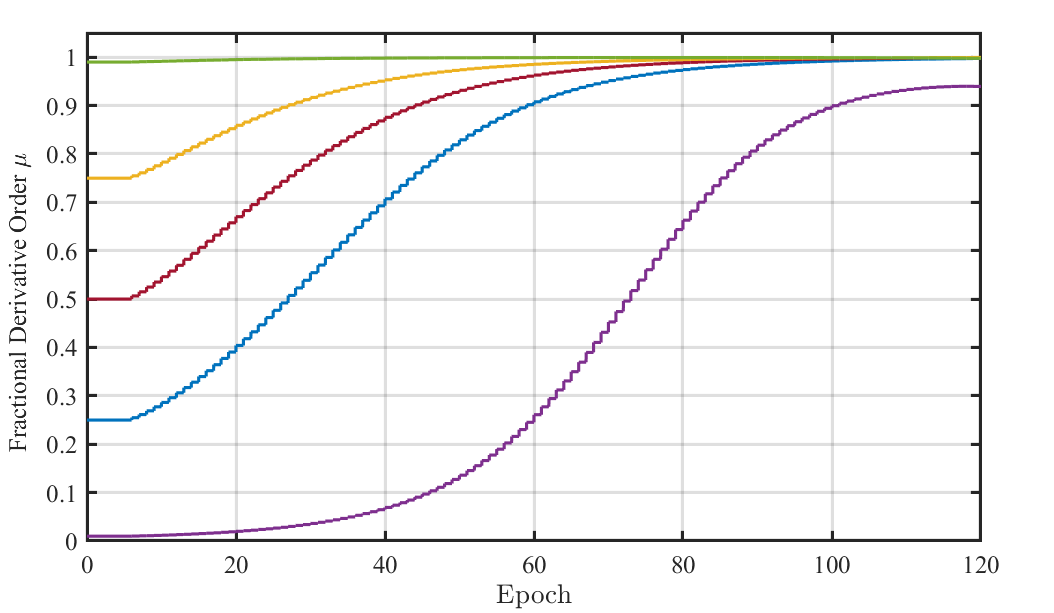}
  \label{fig:initial mu alpha}}
  \caption{(a) Validation accuracy and (b) variation of adaptive \(\mu\) for different initial \(\mu\) values.}
  \label{fig: initial mu ablation}
\end{figure}

In the second ablation study, we investigate the impact of the initial value of \(\mu\) on its learning dynamics and overall model performance. Specifically, we initialize \(\mu\) at \(\{0, 0.25, 0.5, 0.75, 1\}\) and monitor the validation accuracy, as shown in Fig.~\ref{fig:initial mu accuracy}. The evolution of \(\mu\) during training for each initial value is depicted in Fig.~\ref{fig:initial mu alpha}. Since we have a high noise ratio (symmetric \(\eta=0.6\)), we observe that \(\mu\) consistently increases toward 1, reflecting an increased emphasis on robustness (i.e., a shift toward MAE-like behavior). Despite the different starting values, $\mu$ converges to approximately the same value in each case. However, the accuracy curves demonstrate that when \(\mu\) is initialized at 0 or 0.25, the transition to MAE-like behavior is too slow, still resulting in underfitting. In contrast, initializing \(\mu\) at 0.5, 0.75, or 1 prevents overfitting for this experiment. Based on these findings, we conclude that an initial value of \(\mu = 0.5\) is generally a balanced choice, allowing \(\mu\) to adjust effectively according to the noise level. For both lower and higher noise levels, initializing with $\mu(0)=0.5$ remains a reasonable and flexible choice, as it allows the optimizer to adjust $\mu$ upward or downward during training, ultimately converging to a value that reflects the appropriate level of robustness.

\subsubsection{Fixed vs. adaptive \(\mu\)}
\begin{figure}[t]
    \centering
    \includegraphics[width=0.55\linewidth]{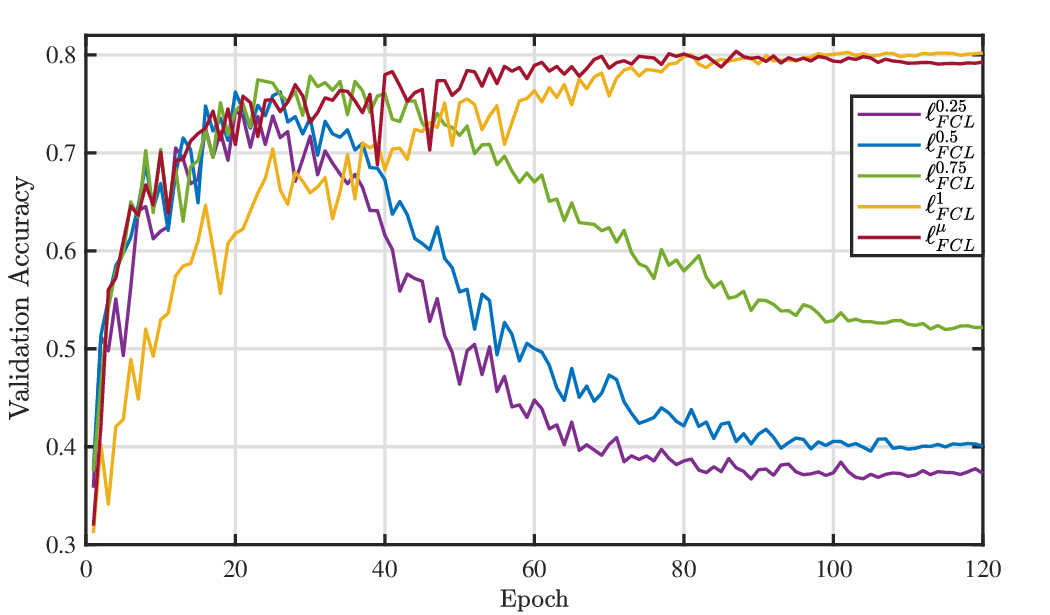}
    \caption{Comparison of FCL performance with fixed versus learnable \(\mu\).}
    \label{fig:different_inits}
\end{figure}

For the third ablation study, we explore the advantages of treating $\mu$ as a learnable parameter instead of a fixed hyperparameter. As indicated in the previous ablation study, under high noise ratios, the optimal strategy is to increase \(\mu\) to enhance robustness. As illustrated in Fig.~\ref{fig:different_inits}, models with fixed \(\mu\) (0.25, 0.5, or 0.75) tend to overfit and underperform relative to the model with a learnable \(\mu\). In this experiment, the best fixed setting is \(\mu = 1\). Here, FCL offers two key advantages: First, the optimal value \(\mu \approx 1\) is not known beforehand and would require expensive hyperparameter tuning, whereas FCL learns it automatically during training. Second, FCL reaches over 70\% validation accuracy much earlier, around 15 epochs compared to 40 epochs for the fixed \(\mu = 1\) model, since at the beginning of the training, FCL is closer to CE-like behavior, enabling faster convergence to high accuracy.

\subsubsection{Time Complexity of FCL}
Table \ref{tab:total_training_times} shows the mean total training times and standard deviations from three CIFAR-10 experiments (0.6 noise ratio) for each loss function. The results show that FCL takes approximately 1.37 times longer than CE, MAE, and GCE and 1.2 times longer than APL losses SCE, NCE+MAE and NCE+RCE. This is because the introduction of the learnable parameter $\mu$ in FCL requires additional gradient calculations, as shown in \eqref{eq: grad w.r.t mu}, which increases computational complexity slightly. Nonetheless, FCL’s ability to dynamically adapt its robustness to label noise, thereby facilitating error minimization while maintaining robustness, represents a fair trade-off given the additional training time.

\subsection{Evaluation on Benchmark Datasets}

\subsubsection{Compared loss functions}
We evaluate our method against: CE, MAE, GCE \cite{zhang2018generalized}, SCE \cite{wang2019symmetric}, and two active-passive losses from \cite{ma2020normalized}: NCE + MAE, NCE + RCE. Our goal is to show that FCL provides robustness to label noise while maintaining state-of-the-art accuracy without any hyperparameter tuning, where $\mu$ is learned during training.

\subsubsection{Noise generation for datasets}
We follow the standard procedures \cite{patrini2017making,ma2018dimensionality} to corrupt training labels with two types of noise. 
Symmetric noise is generated by randomly flipping a certain fraction of labels in each class to other classes. Asymmetric noise is introduced by flipping labels within a subset of classes in a structured manner; for MNIST, flipping $7 \rightarrow 1, 2 \rightarrow 7, 5 \leftrightarrow 6, 3 \rightarrow 8$. For CIFAR-10, we flip \textit{TRUCK} $\to$ \textit{AUTOMOBILE}, \textit{BIRD} $\to$ \textit{AIRPLANE}, \textit{DEER} $\to$ \textit{HORSE}, and \textit{CAT} $\leftrightarrow$ \textit{DOG}. 
For CIFAR-100, we group the 100 classes into 20 super-classes of 5 sub-classes each and flip each class in a circular fashion within its super-class \cite{ma2020normalized}. 
We use the noise rate \(\eta \in \{0.2, 0.4, 0.6, 0.8\}\) for symmetric noise and \(\eta \in\{0.2, 0.4\}\) for asymmetric noise and also give the clean dataset results (i.e. $\eta=0$).
\begin{table}
\centering
\caption{Mean total training times over 3 experiments.}
\label{tab:total_training_times}
\begin{tabular}{cc}
\toprule
\textbf{Loss} & \textbf{Average Total Training Time (s)} \\
\midrule
CE         &  $2369.28 \pm 33.6243$ \\
MAE        &  $2378.92 \pm 21.2295$ \\
GCE  \cite{zhang2018generalized}      &  $2367.14 \pm 57.7509$ \\
SCE    \cite{wang2019symmetric}    &  $2678.47 \pm 111.7834$ \\
NCE+MAE  \cite{ma2020normalized}  &  $2708.31 \pm 97.8374$ \\
NCE+RCE  \cite{ma2020normalized}  &  $2528.89 \pm 41.2839$ \\
\midrule
FCL        &  $3232.64 \pm 76.8291$ \\
\bottomrule
\end{tabular}
\end{table}

\subsubsection{Network architectures and hyperparameters}
We employ models from previous works \cite{ma2020normalized,ye2023active}: a 4-layer CNN for MNIST, an 8-layer CNN for CIFAR-10, and a ResNet-34 for CIFAR-100. The models are trained for 50, 120, and 200 epochs on MNIST, CIFAR-10, and CIFAR-100, respectively. All networks are optimized using the Adam optimizer with a cosine-annealed learning rate schedule and weight decay set to \(1\times 10^{-3}\), \(1\times 10^{-4}\), and \(1\times 10^{-5}\) for MNIST, CIFAR-10, and CIFAR-100, respectively. The initial learning rate is 0.001 for all datasets, whereas the learning rate for learning $\mu$ is 0.1. Standard data augmentation (random shifting and horizontal flipping) is applied, and the gradient norm is clipped to 10 in all settings.

\subsubsection{Parameter Settings}
We adopt the loss hyperparameters recommended in the original papers for each baseline and dataset. For example, we set \(q = 0.7\) for GCE \cite{zhang2018generalized}, and adjust \(\alpha\) and \(\beta\) for SCE, NCE+MAE, and NCE+RCE on a per-dataset basis following \cite{wang2019symmetric, ma2020normalized}. However, as demonstrated in our ablation studies, even slight mis-tuning of these hyperparameters can significantly degrade base loss performance.

\subsubsection{Results and Discussion}
Table~\ref{tab:symmetric_results} reports the mean classification accuracies for 3 experiment runs, obtained under both symmetric ($\eta=\{0.2, 0.4, 0.6, 0.8\}$) and asymmetric label noise ($\eta=0.2, 0.4$). Overall, FCL consistently achieves top performance or second across a range of noise levels and datasets. Notably, under high symmetric noise (e.g., \(\eta=0.6\) or \(0.8\)), FCL outperforms competing methods by a clear margin. For asymmetric noise, FCL also maintains similar robust accuracy. In some cases, APL-based losses achieve slightly better performance without statistical significance due to their hyperparameters being specifically tuned for each dataset; however, as our ablation studies demonstrate, even slight mistuning of these hyperparameters can significantly degrade performance. These results demonstrate the advantage of FCL: by eliminating the need for manual hyperparameter tuning, FCL consistently achieves state-of-the-art results on noisy datasets.
\begin{table}[t!]
\centering
\caption{Mean test accuracies (\%) for different loss functions on benchmark datasets with clean, symmetric, and asymmetric noise.}
\label{tab:symmetric_results}
\begin{adjustbox}{max width=\textwidth}
\begin{threeparttable}
\begin{tabular}{c c c c c c c c c}
\toprule
\multirow{2}{*}{Dataset} 
 & \multirow{2}{*}{Loss } 
 & \multirow{2}{*}{Clean} 
 & \multicolumn{4}{c}{Symmetric Noise Rate ($\eta$)} 
 & \multicolumn{2}{c}{Asymmetric Noise Rate ($\eta$)} \\
\cmidrule(lr){4-7}\cmidrule(lr){8-9}
 &  & & 0.2 & 0.4 & 0.6 & 0.8 & 0.2 & 0.4 \\
\midrule
\multirow{7}{*}{MNIST}
 & CE    & 99.23$\pm$0.05 & 92.61$\pm$0.25 & 84.35$\pm$0.44 & 66.73$\pm$0.96 & 32.93$\pm$1.24 & 94.77$\pm$0.20 & 84.23$\pm$0.71 \\
 & MAE          & 99.15$\pm$0.03 & 98.93$\pm$0.05 & 98.48$\pm$0.03 & 97.14$\pm$0.13 & 72.97$\pm$0.37 & 98.89$\pm$0.06 & 97.15$\pm$0.31 \\
 & GCE \cite{zhang2018generalized}          & 99.11$\pm$0.13 & 98.74$\pm$0.01 & 98.01$\pm$0.15 & 94.63$\pm$0.63 & 71.49$\pm$0.61 & 98.40$\pm$0.06 & 87.63$\pm$0.27 \\
 & SCE \cite{wang2019symmetric}     & 99.07$\pm$0.02 & 98.85$\pm$0.09 & 98.12$\pm$0.11 & 94.55$\pm$0.16 & 72.91$\pm$1.21 & 98.70$\pm$0.06 & 92.88$\pm$0.75 \\
 & NCE+RCE \cite{ma2020normalized}    & 99.06$\pm$0.13 & 98.77$\pm$0.08 & 98.27$\pm$0.09 & 96.40$\pm$0.46 & 81.55$\pm$0.78 & 98.86$\pm$0.16 & 95.28$\pm$0.64 \\
 & NCE+MAE \cite{ma2020normalized}  & 99.35$\pm$0.01 & 99.07$\pm$0.03 & 98.36$\pm$0.17 & 93.92$\pm$0.26 & 81.06$\pm$0.62 & \textbf{99.90$\pm$0.17} & \textbf{95.77$\pm$5.20} \\
 \cmidrule(lr){2-9}
 & \textbf{FCL}    & \textbf{99.97$\pm$0.06} & \textbf{99.93$\pm$0.06} & \textbf{99.90$\pm$0.10} & \textbf{99.47$\pm$0.40} & \textbf{82.70$\pm$1.57} & 97.42$\pm$0.38 & 95.18$\pm$1.05 \\
\midrule
\multirow{7}{*}{CIFAR-10}
 & CE           & 90.69$\pm$0.11 & 76.24$\pm$0.24 & 58.04$\pm$0.33 & 38.74$\pm$1.24 & 25.21$\pm$2.29 & 84.41$\pm$0.46 & 75.01$\pm$0.32 \\
 & MAE          & 91.20$\pm$0.11 & 86.24$\pm$0.41 & 83.89$\pm$0.11 & 77.93$\pm$0.38 & 22.77$\pm$5.03 & 80.59$\pm$0.16 & 58.11$\pm$3.12 \\
 & GCE \cite{zhang2018generalized}          & \textbf{91.87$\pm$0.12} & 89.07$\pm$0.25 & 83.01$\pm$0.08 & 66.06$\pm$0.95 & 36.50$\pm$2.23 & 84.31$\pm$0.64 & 74.84$\pm$0.11 \\
 & SCE \cite{wang2019symmetric}          & 91.16$\pm$0.10 & 86.12$\pm$0.22 & 76.29$\pm$0.48 & 56.47$\pm$0.43 & 27.91$\pm$1.39 & 85.40$\pm$0.72 & 74.01$\pm$0.24 \\
 & NCE+RCE \cite{ma2020normalized}    & 90.71$\pm$0.14 & 88.57$\pm$0.31 & 85.24$\pm$0.25 & 78.29$\pm$0.39 & 45.29$\pm$2.66 & \textbf{87.75$\pm$0.17} & \textbf{76.76$\pm$0.50} \\
 & NCE+MAE \cite{ma2020normalized}    & 91.21$\pm$0.24 & 89.15$\pm$0.31 & 85.38$\pm$0.24 & 78.95$\pm$0.09 & 44.40$\pm$1.70 & 87.69$\pm$0.23 & 76.41$\pm$0.53 \\
\cmidrule(lr){2-9}
 & \textbf{FCL}    & 91.70$\pm$0.13 & \textbf{89.37$\pm$0.15} & \textbf{86.13$\pm$0.24} & \textbf{79.38$\pm$0.14} & \textbf{57.32$\pm$1.43} & 86.47$\pm$0.44 & 75.91$\pm$0.17 \\
 \midrule
\multirow{7}{*}{CIFAR-100}
 & CE           & \textbf{72.89$\pm$0.25} & 57.43$\pm$0.81 & 40.72$\pm$0.45 & 24.04$\pm$1.01 & 7.85$\pm$0.18  & 58.65$\pm$0.12 & 41.43$\pm$0.41 \\
 & MAE          & 11.28 $\pm$1.01           & 7.71 $\pm$0.41           & 8.14 $\pm$0.61           & 4.14 $\pm$1.21           & 2.17 $\pm$1.91           & 6.34 $\pm$0.38          & 3.87 $\pm$0.34 \\
 & GCE \cite{zhang2018generalized} & 66.41$\pm$0.63 & 64.80$\pm$0.20 & 58.99$\pm$0.87 & 44.79$\pm$0.66 & 22.41$\pm$1.05 & \textbf{64.05$\pm$0.63} & 42.64$\pm$0.72 \\
 & SCE \cite{wang2019symmetric}          & 71.76$\pm$0.47 & 56.46$\pm$0.38 & 39.59$\pm$2.14 & 22.29$\pm$0.69 & 7.62$\pm$0.04  & 58.22$\pm$0.41 & 41.50$\pm$0.15\\
 & NCE+RCE \cite{ma2020normalized}    & 72.01$\pm$0.60 & 55.09$\pm$0.34 & 36.45$\pm$2.43 & 19.12$\pm$1.33 & 9.72$\pm$3.27           & 61.95$\pm$0.35 & 39.87 $\pm$1.41 \\
 & NCE+MAE \cite{ma2020normalized}    & 67.86$\pm$0.78 & 62.79$\pm$1.39 & 53.20$\pm$2.15 & 15.21$\pm$24.61 & 5.90$\pm$8.48  & 58.55$\pm$0.56 & 40.20$\pm$1.13 \\
 \cmidrule(lr){2-9}
 & \textbf{FCL}    & 70.59 $\pm$ 0.52   & \textbf{66.94$\pm$0.29} & \textbf{59.15$\pm$0.98} & \textbf{45.19$\pm$1.50} & \textbf{25.91$\pm$0.51}  & 59.73$\pm$0.66 & \textbf{43.12$\pm$0.94} \\
\bottomrule
\end{tabular}
\end{threeparttable}
\end{adjustbox}
\begin{tablenotes}
\item The best results are in bold. \\
\end{tablenotes}
\end{table}

\begin{figure*}[t]
    \centering
    \captionsetup[subfigure]{labelformat=empty} 
    
    \subfloat{%
        \includegraphics[width=\linewidth]{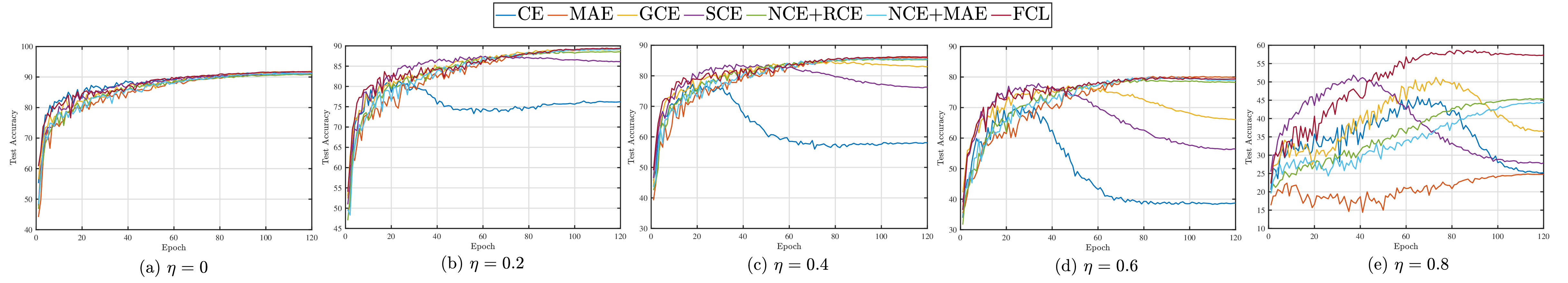}%
    \label{fig:subfig1}}%
    \vspace{-0.01cm}
    \subfloat{%
        \includegraphics[width=\linewidth]{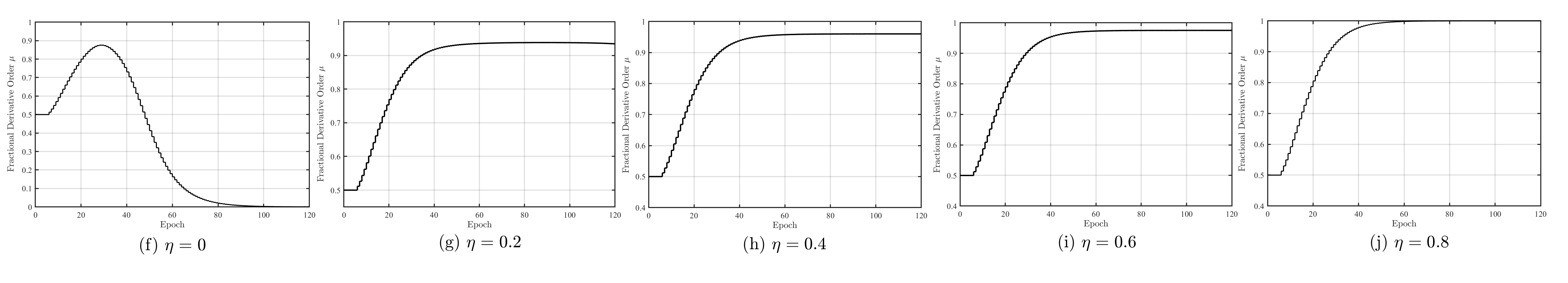}%
    \label{fig:subfig2}}%
    \vspace{-0.1cm}
    \captionsetup[subfigure]{labelformat=parens} 
\caption{Illustrative results from a single run on the CIFAR-10 dataset under symmetric noise levels $\eta = \{0, 0.2, 0.4, 0.6, 0.8\}$, (a)--(e): Test accuracies, (f)--(j): Variation of adaptive $\mu$ for FCL.}
    \label{fig:loss_comparison}
\end{figure*}

For further illustration purposes, Fig.~\ref{fig:loss_comparison}(a)--(e) shows the test accuracy curves for CIFAR-10 under varying noise ratios. It is evident that FCL never underfits, consistently yielding the best or second-best performance without any manual tuning. The variation of adaptive \(\mu\) values, presented in Fig.~\ref{fig:loss_comparison}(f)--(j), indicates that as the noise ratio increases, \(\mu\) tends toward higher values, approaching 1 for noise ratios of 0.6 and 0.8, to achieve MAE-like robustness. Conversely, for a clean dataset (Fig.~\ref{fig:loss_comparison}(f)), \(\mu\) initially rises and then decreases toward zero, reflecting a shift toward CE-like behavior, which is appropriate when label noise is minimal. Thus, \(\mu\) is learned in such a way that it automatically balances robustness and convergence based on the noise level in the dataset.

\section{Conclusion and Future Work}\label{sec: Conclusion and Future Work}
This paper proposed a novel family of adaptive and robust loss functions, called FCL. Leveraging the fractional derivative order $\mu$, FCL morphs its loss landscape between CE-like and MAE-like shapes. Increasing the fractional derivative order \(\mu\) enhances the robustness of the learning process against noisy labels by reducing the penalty on noisy samples, while concurrently increasing the penalty on well-classified examples. Thanks to this unique property, rather than relying on manual tuning of \(\mu\), we integrated it into gradient-based optimization, allowing the model to automatically adjust its robustness level in response to the inherent noise in the dataset. Extensive experiments on benchmark datasets under both symmetric and asymmetric noise conditions reveal that replacing standard loss functions with FCL eliminates robustness hyperparameter tuning and improves classification performance. These results demonstrate the potential of FCL as a powerful tool for robust deep learning in the presence of label noise.

For our future work, we aim to integrate fractional order operators into the temporal-difference error of reinforcement learning algorithms to enhance adversarial robustness. Additionally, we plan to apply these operators within graph neural networks for graph classification and regression tasks, thereby improving robustness against noise in graph data.

\section*{CRediT authorship contribution statement}

    \textbf{Mert Can Kurucu:} Writing – review $\&$ editing, Writing – original draft, Visualization, Validation, Software, Methodology, Investigation, Formal analysis, Data curation. \textbf{Tufan Kumbasar:} Writing – review $\&$ editing, Writing – original draft, Supervision, Methodology, Investigation. \textbf{Ibrahim Eksin:} Writing – review $\&$ editing, Writing – original draft, Supervision. \textbf{Müjde Güzelkaya:} Writing – review $\&$ editing, Writing – original draft, Supervision.

\section*{Acknowledgments}

This work was supported by the Scientific Research Projects Unit of Istanbul Technical University (ITU BAP) under Project No. MDK-2023-44497. The funding source had no involvement in the study design, data collection, analysis, or decision to publish.

During the preparation of this work, the authors used ChatGPT in order to refine the grammar and enhance the English language expressions. After using ChatGPT, the authors reviewed and edited the content as needed and take(s) full responsibility for the content of the publication.


\section*{Appendix}
This appendix presents the fractional derivative of power functions, which is used in the development of fractional loss functions. The classical \(n^\text{th}\) derivative \(\frac{d^n f(x)}{dx^n}\) is well defined for integer \(n\). Fractional calculus extends this notion to non-integer orders. For a function \(f(x)\), its fractional derivative of order \(\mu\) is denoted by
\begin{equation}
    D^\mu f(x) = \frac{d^\mu f(x)}{dx^\mu}, \quad \mu \in \mathbb{R}.
\end{equation}

In this paper, we will use the fractional derivative for power functions \cite{miller1993}. Consider the power function \(f(x) = x^k\). Its standard integer-order derivative is given by
\begin{equation}
    \frac{d^n x^k}{dx^n} = \frac{k!}{(k-n)!}\, x^{k-n}, \quad n \le k.
\end{equation}
To extend this definition to non-integer orders, we replace the factorial with the gamma function, defined as
\begin{equation}
    \Gamma(z) = \int_0^\infty t^{z-1}e^{-t}\, dt,
\end{equation}
which is defined for non-integer $z$ values and also satisfies \(\Gamma(n+1) = n!\) for \(n \in \mathbb{N}\). Consequently, the fractional derivative of \(x^k\) is defined as:
\begin{equation}
    \label{eq: power function FD}
    D^\mu x^k = \frac{\Gamma(k+1)}{\Gamma(k+1-\mu)}\, x^{k-\mu},
\end{equation}
which holds for \(x, k \geq 0\).

An efficient formulation for computing the gamma function is
\begin{equation}
    \label{eq: gamma function}
    \Gamma(z) = \frac{e^{-\gamma z}}{z}\prod_{r=1}^{\infty}\left[\left(1+\frac{z}{r}\right)^{-1}e^{\frac{z}{r}}\right],
\end{equation}
where \(\gamma \approx 0.57721\) is the Euler–Mascheroni constant.

\bibliographystyle{elsarticle-num.bst}
\bibliography{bibliography}

\end{document}